\documentclass{article}

\usepackage[main, final]{neurips_2025}

\usepackage[utf8]{inputenc} 
\usepackage[T1]{fontenc}    
\usepackage{hyperref}       
\usepackage{url}            
\usepackage{booktabs}       
\usepackage{amsfonts}       
\usepackage{nicefrac}       
\usepackage{microtype}      
\usepackage{xcolor}         
\usepackage{amsmath}
\usepackage{multirow}
\usepackage{graphicx}
\usepackage{threeparttable}
\makeatletter
\renewcommand{\maketag@@@}[1]{\hbox{\m@th\normalsize\normalfont#1}}%
\makeatother

\title{GEN: A Practical Alternative to Graph Transformers for Long-Range Graph Modeling}

\author{
  Shuo Wang\textsuperscript{1}, Ge Cheng\textsuperscript{1}\thanks{Corresponding author}, Yun Zhang\textsuperscript{2} \\
  \textsuperscript{1}School of Computer Science \& Cyberspace Science, Xiangtan University, Xiangtan, Hunan, China \\
  \textsuperscript{2}Hunan University, Changsha, Hunan, China \\
  \texttt{202331630305@smail.xtu.edu.cn, chengge@xtu.edu.cn, yunzhangcn@outlook.com}
}

\begin{document}

\maketitle

\begin{abstract}
Message Passing Neural Networks (MPNNs) model local relations effectively but struggle to propagate information over long distances. Graph Transformers (GTs) mitigate this via global self-attention, yet their quadratic cost in the number of nodes limits scalability. We propose Graph Elimination Networks (GENs), an MPNN variant that approximates GT–like long-range modeling while maintaining high efficiency. GENs combine edge-wise and hop-wise self-attention in parallel; their multiplicative composition yields an attention kernel separable across edge and hop factors within a bounded $K$-hop receptive field. To enable hop-wise attention, we introduce the Graph Elimination Algorithm (GEA), which prevents double counting across hops, ensuring that each round injects the $k$-hop incremental contribution exactly once. Taking differences between successive rounds recovers the $k$-hop increment and yields disentangled multi-hop features as inputs for hop-wise attention. This preserves clearer structural distinctions across hop distances and enables more faithful modeling of pairwise dependencies between distant nodes within the $K$-hop neighborhood. On the Long-Range Graph Benchmark (LRGB), GENs outperform strong MPNN baselines by 7.7 and 6.0 percentage points (pp) on PascalVOC-SP and COCO-SP, and achieve performance on par with or better than state-of-the-art Graph Transformers. On OGBN-Products, GENs support full-batch training/inference, while sparse-attention baselines like Exphormer struggle with memory limits under comparable budgets, highlighting GENs as a practical alternative for large, sparse graphs. Our code is available at \href{https://github.com/tmp488598/Graph-Elimination-Networks}{URL}.
\end{abstract}

\section{Introduction}

Message Passing Neural Networks (MPNNs) have become the dominant framework for Graph Neural Networks (GNNs) \cite{ref1}, with wide-ranging applications in social networks \cite{ref2}, recommender systems \cite{ref3}, bioinformatics \cite{ref4,ref5,ref6}, molecular modeling in chemistry \cite{ref7,ref8}, and physical simulation \cite{ref9}. By relying on multi-round local neighbor aggregation, MPNNs achieve high computational efficiency and scalability on large graphs. However, empirical studies have shown \cite{ref57} that their performance often degrades markedly on tasks involving long-range dependencies. Prior work attributes this to several factors, most notably topological bottlenecks (over-squashing) \cite{ref22, ref58, ref114} and over-smoothing \cite{ref16, ref19, ref20}. Beyond these well-documented issues, we draw attention to a factor that has received relatively little attention in recent years.

Specifically, long-range information must traverse multiple propagation layers before reaching a target node, during which signal attenuation \cite{ref18} and multi-hop mixing \cite{2020review, ref67} readily occur, thereby eroding discriminability with respect to non-immediate neighbors. For example, although Graph Attention Networks (GATs) can adaptively weight neighbor features, their attention weights accumulate multiplicatively along paths (e.g., \(\alpha_{ij} \alpha_{jk} \alpha_{kl}\)). Such cross-layer products induce multiplicative dependence on—and attenuation across—the set of paths, making it difficult to explicitly model arbitrary node pairs in a single step. By contrast, Graph Transformers (GTs) \cite{gt2025} use global self-attention to explicitly model interactions between any pair of nodes and therefore excel on long-range tasks, but their computational and memory costs typically scale as \(\mathcal{O}(|V|^2)\), posing a severe burden for large, sparse graphs.

Motivated by these observations, we introduce \emph{hop-wise attention} on top of the \emph{edge-wise (within-hop) attention} paradigm of GATs, yielding a separable attention kernel over a bounded \(K\)-hop neighborhood. Concretely, the affinity between a source and a target node factorizes multiplicatively into (i) an edge-wise term that selects informative neighbors within a fixed hop and (ii) a hop-wise term that modulates contributions by topological distance, producing a parameterization separable across edge and hop factors. Within this design, the key challenge lies in constructing clean, hop-isolated signals that can serve as well-defined inputs for hop-wise attention.

Existing attempts to incorporate multi-hop information largely fall into two categories. (1) Random-walk or diffusion--based methods (e.g., APPNP, MAGNA) diffuse at the kernel level and aggregate all length-\(k\) \emph{walks} within each hop, thereby conflating interactions across hops and allowing short-hop effects to re-enter through longer paths; as a result, exactly-\(k\)-hop signals are hard to isolate and within-hop neighbor selection is obscured. (2) Methods based on parallel multi-order responses or layer/hop aggregation (e.g., MixHop \cite{mixhop}, JKNet \cite{ref67}, DAGNN \cite{ref21}, NAGphormer \cite{ref62}) operate by re-weighting or combining representations only \emph{after} layer-wise propagation has already entangled information across hops. As a result, they struggle to recover clean hop-specific increments and lack any explicit mechanism for edge-wise attention within the same hop.

\begin{figure}[t]
\centering
\includegraphics[width=1.0\columnwidth]{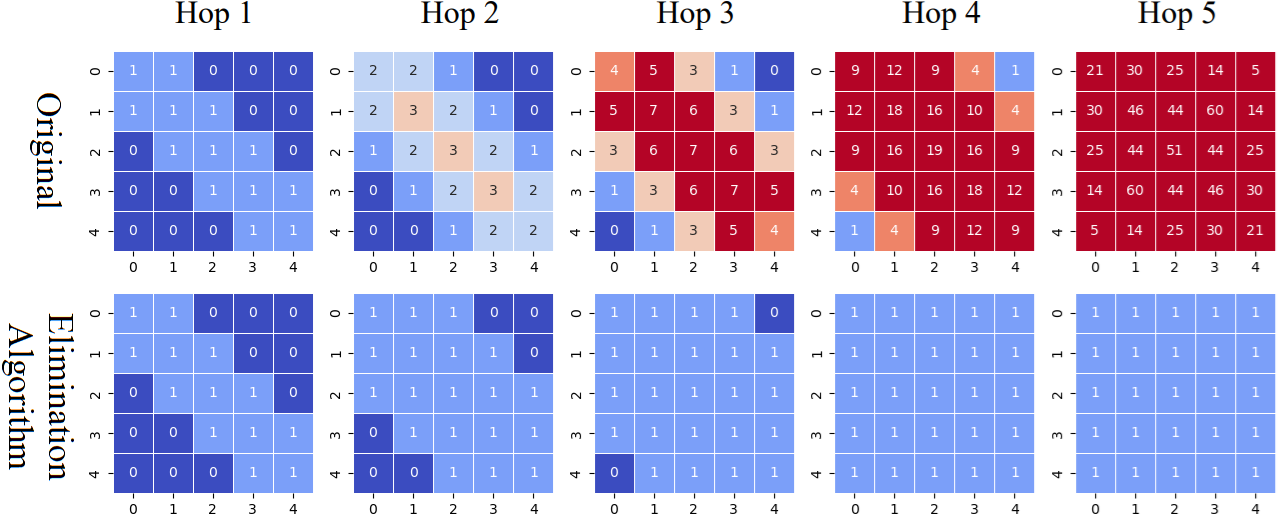}
\caption{Evolution of the feature matrix over five propagation rounds for GENs on a 5-node chain graph. By taking differences between successive rounds, a per-hop decomposition is recovered that disentangles contributions across hop distances. Nodes are initially one-hot encoded, with parameters and normalization omitted for clarity.}
\label{apd:fig1}
\end{figure}

In this paper, we introduce \emph{Graph Elimination Networks} (GENs), an iterative message-passing architecture that performs multiple linear propagation steps within a single network layer. At its core is the \emph{Graph Elimination Algorithm} (GEA), which eliminates cross-hop reuse—and the associated entanglement of multi-hop signals—within GENs’ layer-internal iterative propagation, ensuring that each round injects only the $k$-hop incremental term, as illustrated in Figure~\ref{apd:fig1}. Differencing successive rounds recovers a per-hop decomposition that disentangles contributions across hop distances and provides separable inputs for hop-wise attention Building on this decomposition, GENs apply edge-wise and hop-wise self-attention in parallel: the former selects informative neighbors within the same hop, while the latter attends over topological distances across hops. Their multiplicative composition yields a separable attention kernel within a bounded $K$-hop receptive field, as defined in Eq.~(\ref{eq13b}). The overall time complexity of a single GEN layer is \(\mathcal{O}\!\big(K\,(|E|+|V|)\big)\)\footnote{Throughout, \(K\) denotes the maximum number of propagation steps per layer; \(k\in\{1,\dots,K\}\) indexes the hop distance when we refer to \(k\)-hop neighborhoods; \(|V|\) and \(|E|\) denote the numbers of nodes and edges, respectively.}. 

On the Long-Range Graph Benchmark (LRGB) \cite{ref72}, GENs improve over the strongest MPNN baselines by 7.7 and 6.0 percentage points on PascalVOC-SP and COCO-SP, respectively, achieving performance comparable to or surpassing state-of-the-art GT models. On the large-scale OGBN-Products dataset \cite{ref34}, GENs exhibit computational costs similar to GCNs, whereas sparse-attention methods such as Exphormer exceeded memory limits on our hardware under comparable parameter budgets. These results indicate that GENs effectively model long-range dependencies while maintaining linear scalability. Our contributions are as follows:
\begin{itemize} 
\item{We introduce the GEA, which eliminates cross-hop reuse during layer-internal propagation and prevents double counting, ensuring that each round injects only the $k$-hop incremental term. Differencing successive rounds yields a per-hop decomposition that disentangles hop-specific contributions and provides separable inputs for hop-wise attention.}
\item{We develop GENs as a special case of the MPNN framework. By introducing edge-wise and hop-wise attention in parallel—whose multiplicative combination yields a separable attention mechanism within a local receptive field—GENs strike a balance between efficiency and expressivity.}
\item{We validate the effectiveness and scalability of GENs on the LRGB, large-scale OGBN, and a variety of small- to medium-scale homophilous and heterophilous graph tasks, demonstrating their potential as a practical alternative in sparse regimes.}
\end{itemize}

\section{Related Work}\label{sec2}

\subsection{Message Passing Neural Networks}\label{sec2.1}
Message Passing Neural Networks (MPNNs) \cite{ref13} form the backbone of many graph learning models by iteratively aggregating and propagating information across nodes. Widely adopted variants include GAT \cite{ref11}, GCN \cite{ref10, ref27}, GatedGCN \cite{ref81}, and GIN \cite{ref35}. Despite their success, MPNNs suffer from several fundamental limitations: their expressiveness is bounded by the Weisfeiler-Leman (1-WL) isomorphism test \cite{ref35}, they encounter over-smoothing as depth increases \cite{ref17, ref18, ref82}, and they are vulnerable to over-squashing, which compresses long-range information \cite{ref57, ref61}. To address these issues, efforts have focused on developing higher-order architectures \cite{ref42, ref43}, deeper models \cite{ref25, ref26, ref23, ref22}, adaptive propagation techniques \cite{ref114, ref115}, and graph rewiring strategies \cite{ref83, ref68}. Despite substantial research efforts, the exploration of strategies to mitigate signal attenuation and multi-hop mixing remains relatively limited, leaving room for subsequent methods to further enhance the modeling of long-range information.

\subsection{Graph Transformers}\label{sec2.2}
Graph Transformers (GTs) represent a class of graph learning architectures that incorporate the Transformer’s self-attention with graph-specific inductive biases. Unlike standard transformers for Euclidean or sequential data \cite{ref31}, GTs integrate structural priors through various adaptations. Early approaches such as GPT-GNN \cite{ref108} employ pooling/unpooling to capture multi-scale relations, whereas Graphormer \cite{ref32} encodes global structure via structural/positional and edge encodings (e.g., distance- and centrality-related biases). Subsequent work (e.g., Parker et al. \cite{ref104}) explores SVD-based positional encodings, and GraphiT \cite{ref101, ref102, ref103} introduces local edge biases to better capture fine-grained topology. Hybrid methods, including SAT \cite{ref33}, couple message-passing GNN layers with a global Transformer, while GraphGPS \cite{ref65} provides a modular framework that decouples local operators from Transformer-based global modules. Despite these advances, GTs often incur high computational cost on large graphs. To alleviate this, sparse attention mechanisms such as Exphormer \cite{ref71, ref74} and localized attention methods like NAGphormer \cite{ref62} and VCR-Graphormer \cite{ref113} restrict the attention scope. More recent approaches, including NodeFormer \cite{ref96} and SGFormer \cite{ref95}, adopt linear approximations of global attention. Nevertheless, empirical studies \cite{ref76, ref85} indicate that these strategies may still struggle to fully capture genuine long-range dependencies in some regimes.

\subsection{Graph Mamba}\label{sec2.3}
In pursuit of linear-complexity solutions for large-scale graph tasks, Mamba \cite{ref76} has emerged as a promising architecture that combines a state-space model (SSM) with a selective self-attention mechanism \cite{ref89}. Built on a recurrent formulation \cite{ref88} and enhanced by the HiPPO algorithm \cite{ref90} to approximate historical input trajectories, Mamba reduces the exponential memory decay of traditional RNNs to polynomial complexity. This selective attention helps retain critical information in hidden states, enabling effective memory management akin to transformers. The GMN framework \cite{ref86} offers a general design for Mamba-based GNNs, while Graph-Mamba \cite{ref85} integrates MPNNs with Mamba by replacing the Transformer block in GraphGPS. Additionally, Chen et al. \cite{ref84} show that combining deep random walks with Mamba attains performance competitive with state-of-the-art GTs across various datasets. However, the sequential nature of state updates can limit parallelism during inference (and in certain training settings), and stability/forgetting trade-offs in state-space updates \cite{ref91} warrant further investigation for efficient and stable graph modeling.

\section{Proposed Graph Elimination Algorithm} \label{sec3}

In this section, we introduce the Graph Elimination Algorithm (GEA), which disentangles hop-specific information, thereby making the separation of multi-hop information explicit. At each propagation step, GEA concurrently computes and masks edge-wise contributions irrelevant to the current hop, ensuring that each node assimilates only the new information associated with that hop and thereby explicitly extracts and retains multi-hop structural features. The rest of this section is devoted to the core principles and derivations of GEA.

\subsection{Notations} \label{sec3.1}
We begin by defining the essential notation used throughout this section. Consider an undirected graph \(G = (V, E)\), where \(V\) is the set of nodes and \(E\) is the set of edges. The node feature matrix is \(X \in \mathbb{R}^{|V| \times F}\), with each node \(i \in V\) associated with a feature vector \(X_i\). The neighborhood of node \(i\), denoted by \(\mu(i)\), refers to the set of its direct neighbors, and may include \(i\) itself if self-loops are present. We set \(H^{(0)} = X\).

Within the MPNN framework, we use GAT as a canonical edge-attention instantiation to motivate and discuss GEA. In the GAT setting, where self-loops are explicitly added, computations at layer \(l\) are performed over the neighborhood \(\mu(i)\) (which includes \(i\) itself):
\begin{align}
e_{ij}^{(l)} &= \operatorname{LeakyReLU}\!\Big(\mathbf{a}^{(l)}\!\big[\,W^{(l)} h_i^{(l-1)} \,\|\, W^{(l)} h_j^{(l-1)}\big]\Big),
\end{align}
where \(j \in \mu(i)\). Here, \(W^{(l)}\) is a learnable weight matrix, \(\mathbf{a}^{(l)} \in \mathbb{R}^{2F}\) is a learnable attention vector, \(\|\) denotes concatenation, and \(\operatorname{LeakyReLU}(\cdot)\) is the scoring nonlinearity. The scores \(e_{ij}^{(l)}\) are softmax-normalized to yield the attention coefficients:
\begin{align}
\alpha_{ij}^{(l)} = \frac{\exp\!\big(e_{ij}^{(l)}\big)}
{\sum_{k\in\mu(i)} \exp\!\big(e_{ik}^{(l)}\big)}.
\end{align}
The node update explicitly separates the self term and the neighbor aggregation:
\begin{align}
h_i^{(l)} = \sigma\!\Big(
\alpha_{ii}^{(l)} W^{(l)} h_i^{(l-1)}
+ \sum_{j\in \mu(i)-i} \alpha_{ij}^{(l)} W^{(l)} h_j^{(l-1)}
\Big),
\end{align}
where \(\sigma(\cdot)\) is the post-aggregation activation. 

For clarity, we extract the parameter matrix and rewrite the update rule as:
\begin{align}\label{eq1}
h_i^{(l)} 
= \sigma W^{(l)}\!\Big(
\alpha_{ii}^{(l)} h_i^{(l-1)} 
+ \sum_{j \in \mu(i)-i} \alpha_{ij}^{(l)} h_j^{(l-1)}
\Big).
\end{align}

\subsection{Derivation of the Graph Elimination Algorithm} \label{sec3.2}

As exemplified by Eq.~(\ref{eq1}), edge-attention MPNNs such as GAT iteratively aggregate representations from already visited nodes, thereby entangling information across multiple hops. This occurs because the graph structure inherently contains paths that allow nodes to be revisited, including undirected edges, backtracking edges, and self-loops. The goal of GEA is to disentangle hop-specific features—equivalently, to eliminate the redundancy introduced by such edges during propagation. 

Following this elimination principle, it is straightforward to see that the first GAT layer does not introduce redundancy, since each node only aggregates features from its immediate neighbors. Redundancy begins to appear from the second layer onward. Specifically, at the second layer, the update of node \(i\) is given by:
\begin{equation}\label{eq2}
h_i^{(2)}=\sigma W^{(2)}\Big(\alpha_{ii}^{(2)}h_i^{(1)}+\sum_{j\in\mu(i)-i}\alpha_{ij}^{(2)}h_j^{(1)}\Big).
\end{equation}
We focus on rewriting the second term on the right-hand side in order to extract the redundant 1-hop neighbor features already embedded in $h_i^{(1)}$. Expanding \(h_j^{(1)}\) yields:
\begin{align}\label{eq3}
\sum_{j\in\mu(i)-i}\alpha_{ij}^{(2)}h_j^{(1)}
= \sum_{j\in\mu(i)-i}\alpha_{ij}^{(2)}\sigma W^{(1)}\Big(\alpha_{jj}^{(1)}h_j^{(0)}+\alpha_{ji}^{(1)}h_i^{(0)}+\sum_{k\in\mu(j)-j-i}\alpha_{jk}^{(1)}h_k^{(0)}\Big).
\end{align}

Since $\sigma(\cdot)$ is nonlinear function, direct subtraction of this term to eliminate redundancy is generally infeasible. For discussion, suppose $\sigma(x)=\mathrm{ReLU}(x)=\max(0,x)$, and further assume that all initial features $h^{(0)}$ share the same sign, while parameters within each column of the weight matrix $W$ also maintain consistent signs (though signs may differ across columns). Under these conditions, terms inside the nonlinear function can be separated. It is worth noting that the proposed GENs perform multi-round propagation within a single layer without employing activation functions or trainable weight matrices during propagation, and are therefore not subject to these constraints. Noting that \(\alpha_{ij}^{(l)}\geq 0\), we have:
\begin{footnotesize}
\begin{equation}\label{eq4}
\begin{aligned}
\sum_{j\in\mu(i)-i} \alpha_{ij}^{(2)}h_j^{(1)} 
&=\sum_{j\in\mu(i)-i} \alpha_{ij}^{(2)}W^{(1)}\big(\alpha_{jj}^{(1)}h_j^{(0)}+\alpha_{ji}^{(1)}h_i^{(0)}\big)  + \sum_{j\in\mu(i)-i} \alpha_{ij}^{(2)}\Big|W^{(1)}\big(\alpha_{jj}^{(1)}h_j^{(0)}+\alpha_{ji}^{(1)}h_i^{(0)}\big)\Big|\\
&\quad + \sum_{j\in\mu(i)-i} \alpha_{ij}^{(2)}W^{(1)}\sum_{k\in\mu(j)-j-i} \alpha_{jk}^{(1)}h_k^{(0)}  + \sum_{j\in\mu(i)-i} \alpha_{ij}^{(2)}\Big|W^{(1)}\sum_{k\in\mu(j)-j-i} \alpha_{jk}^{(1)}h_k^{(0)}\Big| \\
&=\sum_{j\in\mu(i)-i} \alpha_{ij}^{(2)}\sigma W^{(1)}\big(\alpha_{jj}^{(1)}h_j^{(0)}+\alpha_{ji}^{(1)}h_i^{(0)}\big) + \sum_{j\in\mu(i)-i} \alpha_{ij}^{(2)}\sigma W^{(1)}\sum_{k\in\mu(j)-j-i} \alpha_{jk}^{(1)}h_k^{(0)}.
\end{aligned}
\end{equation}
\end{footnotesize}

Here, \(h_i^{(1)}\) in Eq.~(\ref{eq2}) already encodes 1-hop features. Thus, Eq.~(\ref{eq4}) contains redundant 0-hop and 1-hop features that must be eliminated. We define this redundant term as:
\begin{equation}\label{eq5}
r_i^{(1)}=\sum_{j\in\mu(i)-i}\alpha_{ij}^{(2)}\sigma W^{(1)}\left(\alpha_{jj}^{(1)}h_j^{(0)}+\alpha_{ji}^{(1)}h_i^{(0)}\right).
\end{equation}
Subtracting this redundancy yields the corrected update:
\begin{equation}\label{eq6}
h_i^{(2)}=\sigma W^{(2)}\Big(\alpha_{ii}^{(2)}h_i^{(1)}-r_i^{(1)}+\sum_{j\in\mu(i)-i}\alpha_{ij}^{(2)}h_j^{(1)}\Big).
\end{equation}
Alternatively, using Eq.~(\ref{eq4}), we obtain an equivalent form:
\begin{footnotesize}
\begin{equation}\label{eq7}
h_i^{(2)}=\sigma W^{(2)}\Big(\alpha_{ii}^{(2)}h_i^{(1)}+\sum_{j\in\mu(i)-i}\alpha_{ij}^{(2)}\sigma W^{(1)}\sum_{k\in\mu(j)-j-i}\alpha_{jk}^{(1)}h_k^{(0)}\Big).
\end{equation}
\end{footnotesize}

This procedure can be extended to three or more layers by iteratively identifying and removing redundancy terms \(r_i^{(2)}, r_i^{(3)}, \dots\). Through successive substitution and expansion of neighborhood sets, one can show that redundancy at any layer can ultimately be expressed in terms of the initial features \(h_i^{(0)}\) and \(h_j^{(0)}\). The detailed derivation is provided in Appendix~\ref{apd1}, where we formalize this process by defining a recursive function.
\begin{footnotesize}
\begin{align}\label{eq8}
&f^{(1)}_{ij}=\alpha_{ij}^{(2)}\sigma W^{(1)}\big(\alpha_{jj}^{(1)}h_j^{(0)}+\alpha_{ji}^{(1)}h_i^{(0)}\big), \nonumber \\
&f^{(2)}_{ij}=\alpha_{ij}^{(3)}\sigma W^{(2)}\Big(\alpha_{jj}^{(2)}h_j^{(1)}+\alpha_{ji}^{(2)}h_i^{(1)}-\alpha_{ji}^{(2)}\sigma W^{(1)}\big(\alpha_{ii}^{(1)}h_i^{(0)}+\alpha_{ij}^{(1)}h_j^{(0)}\big)\Big), \nonumber \\
&\cdots \nonumber \\
&f^{(l-1)}_{ij}=\alpha_{ij}^{(l)}\sigma W^{(l-1)}\Big(\alpha_{jj}^{(l-1)}h_j^{(l-2)}+\alpha_{ji}^{(l-1)}h_i^{(l-2)}-f^{(l-2)}_{ji}\Big),
\end{align}
\end{footnotesize}
with initialization \(f^{(0)}_i=f^{(0)}_j=0\). The redundancy term of layer \(l\) is then given by:
\begin{align}\label{eq9}
r^{(l-1)}_i=\sum_{j\in\mu(i)-i} f^{(l-1)}_{ij}.
\end{align}

We refer to the above redundancy term and its elimination process as GEA. An illustrative example after applying the elimination algorithm is provided in Fig.~\ref{apd:fig1}, and the result has been experimentally validated. Notably, any propagation scheme that can be expressed in the form of Eq.~(\ref{eq1}), such as GCN, APPNP, and MixHop, can be equipped with GEA, indicating a relatively broad scope of applicability.

\begin{figure}[t]
\centering
\includegraphics[width=1.0\columnwidth]{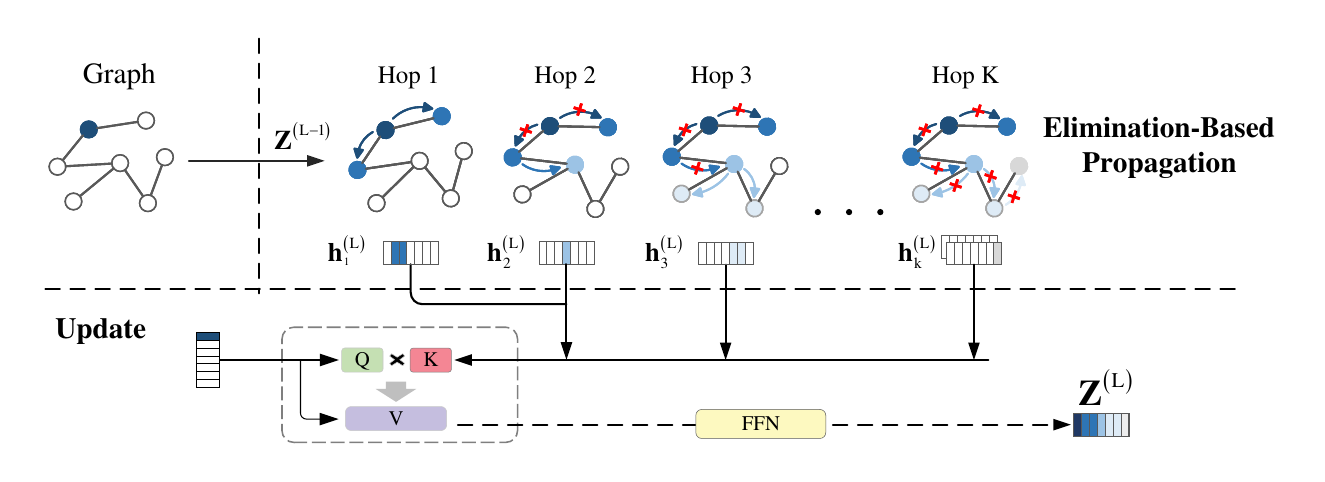}
\caption{Schematic of the $L$-th GEN layer with two stages: \emph{Elimination-Based Propagation} and \emph{Update}. In the propagation stage, edge\mbox{-}wise attention is computed and GEA subtracts redundant contributions to separate representations by hop distance; this is repeated for $K$ rounds. In the update stage, hop\mbox{-}wise self-attention is applied over the $K$ hop-specific representations of node $i$, and the result is passed through a feedforward network (FFN) to produce updated node features, with initial one-hot inputs used only for illustration when specified.}
\label{fig1}
\end{figure}

\section{Methodology}\label{sec4}
Building upon the GEA introduced in Section~\ref{sec3}, we propose Graph Elimination Networks (GENs)—an enhanced architecture grounded in the MPNN framework. GENs introducing a decoupling mechanism \cite{ref21} that separates message passing from feature transformation~\cite{ref28, ref29, ref30}. Specifically, the linear transformation is deferred to the \textit{Update} stage, allowing \textbf{multiple rounds of propagation} within a single layer without involving the activation function \( \sigma \) or the weight matrix \( W \). This design \textbf{preserves the conditions required} by GEA and ensures compatibility with most conventional MPNN variants. We detail the implementation of GENs in two stages: \textit{Elimination-Based Propagation} and \textit{Update}, as illustrated in Figure~\ref{fig1}.

\subsection{Elimination-Based Propagation}\label{sec4.1}

Let \( Z^{(L)} \) denote the node representation matrix updated after the \( L \)-th layer. Define \( h^{(L,k)} \) as the result of the \( k \)-th round of propagation in layer \( L \), with the initial state given by \( h^{(L,0)} = Z^{(L-1)} \). Here, we use \( L \) to represent the current layer of GENs and \( K \) to denote the maximum number of propagation rounds within a single layer. Since the following computations do not involve variables outside layer \( L \), we abbreviate \( h^{(L,k)} \) as \( h^{(k)} \). Based on Eq.~(\ref{eq9}), after ignoring the nonlinear transformation $\sigma W$ (i.e., under decoupling), the computation of redundant terms in the $k$-th round of propagation is given by:
\begin{footnotesize}
\begin{align}\label{eq9a}
f^{(k-1)}_{ij}&=\alpha_{ij}^{\left(k\right)}\big({\alpha_{jj}^{\left(k-1\right)}h}_j^{\left(k-2\right)}+{\alpha_{ji}^{\left(k-1\right)}h}_i^{\left(k-2\right)}-f^{(k-2)}_{ji}\big), \nonumber \\
r^{(k-1)}_{i}&=\sum_{j\in\mu(i)-i}f^{(k-1)}_{ij},
\end{align}
\end{footnotesize}
where \( f^{(0)}_{ij}=f^{(0)}_{ji} = 0 \). It is important to note that \( \alpha_{ij}^{(k)} \) can take any real number, allowing GENs to be compatible with GCN, GAT, or other GNNs. 

Typically, we obtain it by computing edge-wise attention, which, when combined with hop-wise attention in the update stage, forms a dual self-attention selection mechanism that simulates the application of GT methods. The computation for edge-wise attention is given by:
\begin{align}\label{eq10}
e_{ij}^{(k)}\ &=\ {\rm LeakyReLU}\left(\mathbf{a}^{(k)}\left[h_i^{\left(k-1\right)}||h_j^{\left(k-1\right)}\right]\right),\nonumber \\
\alpha_{ij}^{(k)}\ &=\ {\rm softmax}\left(e_{ij}^{(k)}\right),
\end{align}
with \( \mathbf{a}^{(k)} \in \mathbb{R}^{2F} \) denoting the learnable parameter used to compute the edge attention score. The elimination-based propagation stage is defined as:
\begin{align}\label{eq11} 
h_i^{(k)} = \alpha_{ii}^{(k)} h_i^{(k-1)} - r^{(k-1)}_{i} + \sum_{j \in \mu(i)-i} \alpha_{ij}^{(k)}  h_j^{(k-1)} . 
\end{align} 

To balance the scale of features from different hops, we apply a norm-based power compression to each hop representation:
\begin{align}\label{eq11a}
\tilde{h}_i^{(k)} = \frac{h_i^{(k)}}{\big(\lVert h_i^{(k)} \rVert_2 + \varepsilon \big)^{\gamma}}, \quad 0 \leq \gamma \leq 1,
\end{align}
where $\varepsilon$ is a small constant to avoid division by zero and $\gamma$ controls the compression strength. This scaling preserves the direction of each vector while compressing magnitude disparities across hops. 
After completing \( K \) rounds of propagation, the node representations at different distances are combined into a matrix:
\begin{align}\label{eq12} 
H_i^{(L)} = \operatorname{concat}\left( \tilde{h}_i^{(1)}, \tilde{h}_i^{(2)}, \dots, \tilde{h}_i^{(K)} \right) \in \mathbb{R}^{K \times F}, 
\end{align}
which contains the final node representations from all rounds of propagation. All previous equations ignore the \( L \)-th layer for simplicity.

\subsection{Update}\label{sec4.2}

The update layer in GENs is relatively simple and primarily serves to transform the results of multiple rounds of propagation into the output of the hidden layer. \(H_i^{(L)}\) contains the aggregated representations of node \(i\) at different hops. We learn hop-wise attention through a Transformer encoder, allowing node \(i\) to autonomously select the importance of neighbors at different hops. The self-attention module projects the node features and \(H_i^{(L)}\) into three subspaces, \(\mathbf{Q}\), \(\mathbf{K}\), and \(\mathbf{V}\), and then computes:
\begin{align}\label{eq13}
\mathbf{Q}_i &= Z_i^{(L-1)}W^{Q},\quad
\mathbf{K}_i = H_i^{(L)}W^{K},\quad
\mathbf{V}_i = H_i^{(L)}W^{V},
\end{align}
in which \(W^{Q}\), \(W^{K}\), and \(W^{V}\) are learnable parameter matrices. Hop-wise attention captures the affinity between node \(i\) and nodes at different distances within its receptive field. We instantiate the hop-wise attention distribution as:
\begin{align}\label{eq13a}
\beta_{i}
= \operatorname{softmax}\!\left(\frac{\mathbf{Q}_i\mathbf{K}_i^{\top}}{\sqrt{d_{key}}}\right),
\end{align}
where \(\beta_{i}\in\mathbb{R}^{K}\) assigns a probability to each hop layer \(1,\dots,K\) for node \(i\), and \(d_{key}\) is the key dimensionality.

The hop-wise attention adaptively selects the relevant hop layer, and the edge-wise attention adaptively selects the paths to each neighbor in that layer. The resulting composite attention kernel from node \(i\) to any node \(n\) in its receptive field is:
\begin{align}\label{eq13b}
\mathcal{K}_{i,n}
= \prod_{(u,v)\in P_{i\rightsquigarrow n}} \alpha_{uv}\,\beta_{i,\,h(n)},
\end{align}
where \(P_{i\rightsquigarrow n}\) denotes any path from \(i\) to \(n\) within the receptive field, \(\alpha_{uv}\) is the edge-wise attention weight on edge \((u,v)\), and \(h(n)\) is the hop distance from \(i\) to \(n\). 

Ideally, the attention kernel $\mathcal{K}_{i,n}$ can precisely select any neighbor within the receptive field. Consequently, dual attention implicitly enables direct information exchange between node $i$ and any node in its receptive field, without introducing quadratic complexity. If concerns arise regarding the expressivity of the composite attention kernel, it can be enhanced through a multi-head mechanism, where multiple heads learn complementary hop–path kernels that are concatenated and linearly projected. For brevity, the multi-head formulation is omitted here.

Finally, after a residual connection, the result of the dual self-attention is fed into a feed-forward network (FFN), yielding the updated latent space representation:
\begin{align}\label{eq14}
Z_i^{(L)}
= \mathrm{FFN}\!\left(
Z_i^{(L-1)}W^{(L)}
+ \beta_{i}\mathbf{V}_i
\right),
\end{align}
where \(W^{(L)}\) is the learnable parameter of the \(L\)-th layer.

\subsection{Performance Analysis}\label{sec4.3}

We analyze GENs from two perspectives---time complexity and topological embedding---to highlight their efficiency, scalability, and practical advantages, while noting a bounded but mitigable limitation. 

\textbf{Time Complexity.} For MPNN-based methods, the standard complexity is \( \mathcal{O}(|E|+|V|) \), where \( |E| \) and \( |V| \) denote the numbers of edges and nodes. In GENs, the additional cost arises from GEA. As implied by Eq.~(\ref{eq9})--(\ref{eq10}), the elimination terms are computed in parallel with message passing. Across all nodes, the number of neighbor interactions is proportional to \( |E| \); with attention coefficients \( \alpha_{ij} \), feature dimension \( F \), and at most \( K \) propagation rounds, this leads to \( \mathcal{O}(K|E|F) \), which reduces to \( \mathcal{O}(|E|) \) when \( K \) and \( F \) are fixed. The edge-wise attention specified by Eq.~(\ref{eq13})--(\ref{eq14}) incurs \( \mathcal{O}(|V|F^{2}+K|V|F) \), i.e., \( \mathcal{O}(|V|) \) under the same assumption. Therefore, the overall complexity of GENs is \( \mathcal{O}(K(|E|+|V|)) \), which, for constant depth \( K \), matches \( \mathcal{O}(|E|+|V|) \). On sparse graphs where \( |E| \ll |V|^{2} \), GENs are considerably more efficient than standard GTs.  

\textbf{Graph Topological Information.} Effective GNNs must respect permutation symmetries~\cite{ref109}; in practice, this entails permutation equivariance at the node level and invariance after graph-level readout. Efficiently embedding topology, however, remains a nontrivial challenge. Recent GTs~\cite{ref65} often estimate structure using traditional GNNs or random walks and then employ it as positional information for global self-attention, which introduces additional preprocessing and runtime overhead. By inheriting the MPNN framework, GENs can function without this step, since positional cues are implicitly encoded within each node’s receptive-field subgraph~\cite{ref35, ref44, ref45}. Concretely, if \( \alpha_{ij} \) denotes edge-wise attention over 1-hop neighbors, the effective weight on a 2-hop neighbor becomes \( \alpha_{ij}\alpha_{jk} \), which naturally decays with distance because attention weights are normalized to be at most 1. This implicit, distance-aware embedding reduces the need for explicit graph-structure preprocessing while enabling GENs to maintain strong empirical performance and scalability across diverse graph regimes. Nonetheless, incorporating explicit structural information may still offer additional benefits.

\textbf{Limitations and Remedies.} Despite the aforementioned advantages, in the message-passing paradigm, features that reappear along cyclic paths at the same hop distance are indistinguishable from those contributed by new neighbors, making certain topological structures inherently indistinguishable. As a result, MPNN-style models—including GENs with GEA—still share the expressive upper bound of the 1-WL test~\cite{ref35}. In practice, this limitation can be mitigated by incorporating common positional encodings from GTs. Random-walk-based RWSE~\cite{RWSE} and Laplacian eigenvector–based LapPE~\cite{LapPE} have been empirically shown to improve the performance of GENs and other MPNN variants on long-range graph benchmarks rich in cycles. Importantly, these enhancements can be integrated as lightweight plug-ins without undermining GENs’ core efficiency or their ability to operate without graph-structure preprocessing.

\section{Experiments}\label{sec5}

In this section, we evaluate the performance of GENs on the Long-Range Graph Benchmark and the Open Graph Benchmark to assess their ability to model long-term dependencies and to demonstrate their computational advantages on sparse graphs. We further analyze the impact of different choices of \(K\) on runtime and memory consumption to provide a more comprehensive assessment of their cost-effectiveness.

\subsection{Experimental Setup}\label{sec5.1}


\textbf{Datasets and Models.} We evaluated GENs on two primary task categories: (1) long-range dependencies (Long-Range Graph Benchmark, LRGB~\cite{ref72}) and (2) large-scale tasks (Open Graph Benchmark, OGB~\cite{ref34}). We compared GENs against the following baselines: (1) classic MPNN methods~\cite{ref73}, such as GCN~\cite{ref10}, GIN~\cite{ref35}, GAT~\cite{ref11}, Mix-Hop \cite{mixhop}, DAGNN \cite{ref21}, MAGNA \cite{ref80},  GatedGCN \cite{ref81}, IPR-MPNN~\cite{ref112}, and Gated-GCN~\cite{ref81}; (2) state-of-the-art Graph Transformers, including SAT~\cite{ref33}, NAGphormer~\cite{ref62}, GRIT~\cite{ref93}, Drew~\cite{ref83}, GPS~\cite{ref65}, Exphormer~\cite{ref71}; (3) recent Mamba-based approaches, such as Graph-Mamba~\cite{ref85} and GMN~\cite{ref86}. 

\textbf{Settings.} For each benchmark, we follow the official evaluation protocols. For \textit{LRGB}, we adhere to the 500K parameter budget constraint and adopt evaluation metrics consistent with prior work. For \textit{OGB}, we use the official dataset splits and convert graphs to a Compressed Sparse Row (CSR) adjacency format using PyTorch Geometric’s~\cite{pyg} \texttt{ToSparseTensor} transform to enable full-batch training. Detailed dataset statistics and metric definitions are reported in the corresponding result tables. For all tasks, we report the mean and standard deviation of test performance obtained from the model checkpoint that achieves the best validation score. All experiments are conducted on a single NVIDIA A100 80 GB GPU. Additional experiments on small-scale graphs and heterophilous graphs, along with their results and settings, are provided in Appendix~\ref{apd:expde}.

\begin{table}[b]
\centering
\caption{Test performance on Long-Range Graph Benchmark. The best result is highlighted in bold, and the second and third best results are underlined.}
\label{tab1}
\resizebox{\textwidth}{!}{%
\begin{tabular}{lcccccc}
\toprule
\textbf{Dataset} & \textbf{PascalVOC-SP} & \textbf{COCO-SP} & \textbf{Peptides-Func} & \textbf{Peptides-Struct} & \textbf{PCQM-Contact} \\
\midrule
\# Graphs       & 11.4K                 & 123.3K           & 15.5K                  & 15.5K                    & 529.4K                \\
Avg. \# Nodes   & 479.4                 & 476.9            & 150.9                  & 150.9                    & 30.1                  \\
Avg. \# Edges   & 2,710.5               & 2,693.7          & 307.3                  & 307.3                    & 61.0                  \\
Metric & F1 ↑ & F1 ↑ & AP ↑ & MAE ↓ & MRR ↑ \\
\midrule
\textbf{GCN}    & 0.2078 $\pm$ 0.0031 & 0.1338 $\pm$ 0.0007 & 0.6860 $\pm$ 0.0050 &  \underline{0.2460 $\pm$ 0.0007} & 0.3424 $\pm$ 0.0007 \\
\textbf{GAT}       & 0.3335 $\pm$ 0.0045 & 0.2697 $\pm$ 0.0011 & 0.6924 $\pm$ 0.0044 & 0.2530 $\pm$ 0.0010 & 0.3441 $\pm$ 0.0008 \\
\textbf{MixHop}       & 0.3421 $\pm$ 0.0056 & 0.2426 $\pm$ 0.0021 & 0.6920 $\pm$ 0.0037 & 0.2510 $\pm$ 0.0008 & 0.3414 $\pm$ 0.0006 \\
\textbf{DAGNN}       & 0.3613 $\pm$ 0.0084 & 0.2602 $\pm$ 0.0023 & 0.7021 $\pm$ 0.0075 & 0.2645 $\pm$ 0.0023
& 0.3287 $\pm$ 0.0020 \\
\textbf{MAGNA}       & 0.3585 $\pm$ 0.0052 & 0.2662 $\pm$ 0.0016 & 0.6541 $\pm$ 0.0038 & 0.2509 $\pm$ 0.0008 & 0.3364 $\pm$ 0.0008 \\
\textbf{GatedGCN}   & 0.3880 $\pm$ 0.0040 & 0.2922 $\pm$ 0.0018 & 0.6765 $\pm$ 0.0047 & 0.2477 $\pm$ 0.0009 & 0.3495 $\pm$ 0.0010\\
\textbf{IPR-MPNN} &-&-& \textbf{0.7210 $\pm$ 0.0039} & 0.2462 $\pm$ 0.0007 & \underline{0.3516 $\pm$ 0.0102} \\
\midrule
\textbf{SAT}            & 0.3230 $\pm$ 0.0039 & 0.2592 $\pm$ 0.0158 & 0.6384 $\pm$ 0.0121                 & 0.2683 $\pm$ 0.0043 & - \\
\textbf{NAGphormer} & 0.4006 $\pm$ 0.0061 & 0.3458 $\pm$ 0.0070 &-&-&-\\
\textbf{GRIT} &-&-&0.6988 $\pm$ 0.0082 & \underline{0.2460 $\pm$ 0.0012}&-\\
\textbf{Drew} &0.3314 $\pm$ 0.0024& - &\underline{0.7150 $\pm$ 0.0044}& 0.2536 $\pm$ 0.0015 &0.3444 $\pm$ 0.0017\\
\textbf{GPS}           & \underline{0.4440 $\pm$ 0.0065} & \underline{0.3884 $\pm$ 0.0055} & 0.6534 $\pm$ 0.0091 & 0.2509 $\pm$ 0.0010 & 0.3498 $\pm$ 0.0005  \\
\textbf{Exphormer}     & 0.3975 $\pm$ 0.0037 & 0.3455 $\pm$ 0.0009 & 0.6527 $\pm$ 0.0043 & 0.2484 $\pm$ 0.0012 & \textbf{0.3637 $\pm$ 0.0020} \\
\textbf{Graph-Mamba}  & 0.4191 $\pm$ 0.0126 & \underline{0.3960 $\pm$ 0.0175} & 0.6739 $\pm$ 0.0087 & 0.2478 $\pm$ 0.0016 &  0.3395 $\pm$ 0.0013 \\
\textbf{GMN}  & \underline{0.4393 $\pm$ 0.0112} & \textbf{0.3974 $\pm$ 0.0101} &  0.7071 $\pm$ 0.0083 & 0.2473 $\pm$ 0.0025 &- \\
\midrule
\textbf{GENs}& \textbf{0.4653 $\pm$ 0.0097}  & 0.3523 $\pm$ 0.0057& \underline{0.7142 $\pm$ 0.0071}& \textbf{0.2430 $\pm$ 0.0013} & \underline{0.3520 $\pm$ 0.0010}\\
\bottomrule
\end{tabular}
}
\end{table}

\begin{table}[t]
\centering
\caption{Results of the ablation study on the Long-Range Graph Benchmark, evaluating the effects of positional encodings, edge-wise attention (EA), hop-wise attention (HA), and the graph elimination algorithm (GEA). The best result is shown in bold.}
\label{tab5}
\resizebox{\textwidth}{!}{%
\begin{tabular}{lcccccc}
\toprule
\multirow{2}{*}{\textbf{Dataset}} & \textbf{PascalVOC-SP} & \textbf{COCO-SP} & \textbf{Peptides-Func} & \textbf{Peptides-Struct} & \textbf{PCQM-Contact} \\
 & F1 ↑ & F1 ↑ & AP ↑ & MAE ↓ & MRR ↑ \\
\midrule
\textbf{GENs}& 0.4427 $\pm$ 0.0131 & 0.3415 $\pm$ 0.0046 & 0.6986 $\pm$ 0.0079 & 0.2490  $\pm$ 0.0016  & 0.3471 $\pm$ 0.0007 \\
\textbf{GENs+RWSE}& \textbf{0.4653 $\pm$ 0.0097} & 0.3404 $\pm$ 0.0042 & 0.7087 $\pm$ 0.0103 & 0.2521 $\pm$ 0.0010 & 0.3440 $\pm$ 0.0008 \\
\textbf{GENs+LapPE}   & 0.4523 $\pm$ 0.0162 & \textbf{0.3523 $\pm$ 0.0057} & \textbf{0.7142 $\pm$ 0.0071} & \textbf{0.2430 $\pm$ 0.0013} & \textbf{0.3520 $\pm$ 0.0010}\\
\midrule
\textbf{GEN w/o HA} & 0.4055 $\pm$ 0.0124 & 0.2987 $\pm$ 0.0033 & 0.6847 $\pm$ 0.0089 & 0.2472 $\pm$ 0.0009 & 0.3480 $\pm$  0.0007 \\
\textbf{GEN w/o EA}& 0.3694 $\pm$ 0.0144 & 0.3044 $\pm$ 0.0054 & 0.6759 $\pm$ 0.0104 & 0.2438 $\pm$ 0.0010 & 0.3498 $\pm$ 0.0005  \\
\textbf{GENs w/o GEA} & 0.3905 $\pm$ 0.0132 & 0.2969 $\pm$ 0.0048 &0.6889 $\pm$ 0.0092&0.2432 $\pm$ 0.0013& 0.3496 $\pm$ 0.0016\\

\bottomrule
\end{tabular}
}
\end{table}

\begin{table}[b]
\centering
\caption{Efficiency and test performance on OGBN datasets under full-batch training. Reported metrics are time per epoch, peak GPU memory, and test accuracy (\%). Values are mean $\pm$ standard deviation across 10 runs. OOM denotes out-of-memory. Best results are in \textbf{bold}.}
\label{tab3}

\begin{threeparttable}
\centering
\resizebox{1.0\textwidth}{!}{%
\begin{tabular}{lcccccc}
\toprule
\textbf{Dataset} & \textbf{Metric} & \textbf{GCN} & \textbf{NAGphormer} & \textbf{Exphormer} & \textbf{Graph-Mamba} & \textbf{Ours} \\
\midrule
\textbf{OGBN-Arxiv} & Accuracy & 71.74$\pm$0.29 & 71.52$\pm$0.24 & 72.44$\pm$0.28 & 71.78$\pm$0.26 & \textbf{72.76$\pm$0.19} \\
169 K \# N & Train Time (s) & \textbf{0.10} & 1.09 & 1.20 & 11.90 & 0.20 \\
1,166 K \# E & Mem (GB) & \textbf{2.73} & 11.40 & 31.89 & 5.95 & 5.78 \\
\midrule
\textbf{OGBN-Products} & Accuracy & 75.64$\pm$0.21 & OOM & OOM & 74.77$\pm$0.45\tnote{*} & \textbf{79.44$\pm$0.32} \\
2.45 M \# N  & Train Time (s) & \textbf{3.12} & OOM & OOM & 736.00 & 6.09 \\
61.86 M \# E & Mem (GB) & \textbf{41.27} & OOM & OOM & 63.55 & 58.13 \\
\bottomrule
\end{tabular}%
}
\begin{tablenotes}[flushleft]
\scriptsize
\item[*] High training cost on OGBN-Products; performance for reference only (no extensive hyperparameter tuning).
\end{tablenotes}
\end{threeparttable}
\end{table}

\begin{table}[t]
  \centering
  \caption{Effect of depth ($L$) and hops per layer ($K$) on \textsc{OGBN}-Arxiv accuracy (\%) for GENs. Results are mean $\pm$ standard deviation across 10 runs. For each fixed $L$, \textbf{bold} denotes the highest accuracy across $K$.}
  \label{tab:lk_arxiv}
  \resizebox{0.95\textwidth}{!}{%
  \begin{tabular}{@{}cc*{6}{c}@{}}
    \toprule
    \multirow{2}{*}{$L$} & \multirow{2}{*}{Metric} & \multicolumn{6}{c}{OGBN-Arxiv} \\ 
    \cmidrule(lr){3-8}
                         &                         & $K{=}1$ & $K{=}2$ & $K{=}4$ & $K{=}6$ & $K{=}8$ & $K{=}10$ \\
    \midrule
    \multirow{3}{*}{1}
      & Accuracy    & 58.28$\pm$0.07 & 63.93$\pm$0.06 & 66.86$\pm$0.06 & 67.69$\pm$0.07 & 67.97$\pm$0.07 & \textbf{68.06$\pm$0.09} \\
      & Train time (s)  & 0.03 & 0.03 & 0.06 & 0.08 & 0.09 & 0.11 \\
      & Mem (GB)     & 1.89 & 2.05 & 3.38 & 4.79 & 5.78 & 8.59 \\
    \midrule
    \multirow{3}{*}{2}
      & Accuracy   & 69.07$\pm$0.26 & 71.81$\pm$0.16 & 72.34$\pm$0.22 & \textbf{72.47$\pm$0.26} & 72.33$\pm$0.24 & 72.02$\pm$0.23 \\
      & Train time (s)  & 0.06 & 0.08 & 0.12 & 0.17 & 0.23 & 0.27 \\
      & Mem (GB)     & 2.97 & 3.56 & 4.62 & 5.95 & 8.02 & 10.42 \\
    \midrule
    \multirow{3}{*}{3}
      & Accuracy   & 70.79$\pm$0.30 & 72.48$\pm$0.22 & \textbf{72.76$\pm$0.19} & 72.58$\pm$0.22 & 72.27$\pm$0.35 & 71.93$\pm$0.21 \\
      & Train time (s)  & 0.08 & 0.12 & 0.20 & 0.27 & 0.35 & 0.44 \\
      & Mem (GB)     & 3.56 & 4.06 & 5.78 & 6.78 & 8.77 & 11.33 \\
    \midrule
    \multirow{3}{*}{4}
      & Accuracy   & 71.31$\pm$0.23 & 72.46$\pm$0.18 & \textbf{72.55$\pm$0.33}  & 72.21$\pm$0.37 & 71.51$\pm$0.35 & 71.10$\pm$0.57 \\
      & Train time (s)  & 0.11 & 0.15 & 0.26 & 0.38 & 0.48 & 0.59 \\
      & Mem (GB)     & 4.14 & 4.89 & 6.94 & 9.44 & 10.68 & 14.64 \\
    \midrule
    \multirow{3}{*}{5}
      & Accuracy   & 71.48$\pm$0.17 & \textbf{72.29$\pm$0.31} & 72.18$\pm$0.35 & 71.71$\pm$0.46 & 71.04$\pm$0.53 & 70.40$\pm$0.83 \\
      & Train time (s)  & 0.14 & 0.20 & 0.33 & 0.47 & 0.62 & 0.75 \\
      & Mem (GB)     & 4.73 & 5.56 & 7.69 & 10.52 & 13.41 & 15.55 \\
    \bottomrule
  \end{tabular}
}
\end{table}

\subsection{Performance on Long-Range Dependency}\label{sec5.2}

We first evaluate GENs’ ability to capture long-range dependencies using the widely adopted LRGB datasets. As shown in Table \ref{tab1}, across five datasets, GENs with dual self-attention selection mechanisms achieve performance comparable to state-of-the-art Mamba-based and GT models, securing the best results on two tasks and the second-best on two others. On the PascalVOC-SP and COCO-SP tasks, GENs outperform the best traditional GNNs by 7.7 and 6.0 percentage points (pp), respectively. Notably, on these two datasets, traditional GNNs lag significantly behind the best-performing GT models, highlighting the effectiveness of dual self-attention in synthesizing global attention.

To evaluate the contribution of each core component, we conducted an ablation study on the LRGB datasets, with results summarized in Table~\ref{tab5}. Positional encodings widely used in GTs, such as RWSE and LapPE, consistently yield around 1 pp improvement for GENs, indicating that these handcrafted schemes remain beneficial for GNNs. In contrast, removing the GEA module results in substantial drops of 7.5 pp and 5.5 pp on PascalVOC-SP and COCO-SP, respectively, bringing performance close to that of conventional GNNs. This suggests that, without elimination, the dual self-attention mechanism reduces to standard message passing, whereas GEA is essential for synthesizing global attention through bilinear interactions. On the other three datasets, the impact of removing GEA is minor, consistent with the smaller performance gap typically observed between GTs and GNNs on these tasks.

\subsection{Cost Efficiency Evaluation}\label{sec5.4}

To evaluate the computational efficiency of GENs, we conduct full-batch training on large-scale OGBN graphs and compare against state-of-the-art baselines, including NAGphormer, Exphormer, and Graph-Mamba, under a parameter budget of approximately 300K, as reported in Table~\ref{tab3}. The results show that NAGphormer (with local attention) and Exphormer (with sparse attention) do not support full-batch training on the OGBN-Products dataset, and both exhibit substantial gaps in scalability and computational efficiency compared with GCN and Graph-Mamba on large-scale graphs. Graph-Mamba, while featuring linear memory scaling similar to sparse GCN, incurs excessive runtime on OGBN-Products under full-batch training---approximately 200 times slower per epoch than standard GCN. In contrast, GENs maintain comparable memory efficiency to GCN and require only about twice its runtime, while consistently delivering superior performance. These results further demonstrate that GENs strike an effective balance between computational cost and predictive accuracy, making them a practical choice for large-scale graph learning.

On the other hand, we further examined training time, memory consumption, and predictive performance on OGBN-Arxiv as functions of the network depth \(L\) and the number of propagation rounds per layer \(K\). The results are summarized in Table~\ref{tab:lk_arxiv}, where, for each fixed \(L\), the best-performing \(K\) is boldfaced. We observe that, irrespective of depth, the optimal performance on OGBN-Arxiv consistently occurs when the effective receptive field lies in the range of 10--16 hops. This suggests that each dataset may admit a characteristic receptive-field size, and that optimal performance is typically achieved by appropriately distributing receptive-field depth between \(L\) and \(K\). Moreover, while enlarging \(K\) leaves the number of learnable parameters nearly unchanged, it increases memory consumption, indicating that indiscriminate escalation of \(K\) is undesirable. Nevertheless, GENs maintain training time and memory footprints comparable to the GCN baseline while exhibiting consistent accuracy gains, further reinforcing their cost-effectiveness for scalable graph learning.

\section{Conclusion}

In this paper, we propose GENs, a framework that integrates the key ideas of Graph Transformers into the MPNN paradigm. At the core of GENs lies the GEA, which disentangles multi-hop information and enables clean hop-wise attention. Ablation studies demonstrate that this disentangling process is critical for effectively modeling complex long-range dependencies. Building upon GEA, GENs incorporate a dual self-attention selection mechanism that combines edge-wise and hop-wise attention, thereby enabling efficient and adaptive modeling of long-range interactions. This design alleviates the high computational cost of Graph Transformers while addressing the limitations of traditional GNNs in capturing distant dependencies. Empirical results indicate that, relative to Graph Transformers, GENs achieve comparable performance on long-range benchmarks and on small- to medium-scale homophilous graphs, deliver moderate performance gains on heterophilous graphs, and offer substantial efficiency improvements on large-scale tasks. Taken together, these contributions highlight GENs as a scalable and efficient solution for graph-based machine learning, offering both theoretical insights and practical benefits, and pointing to promising directions for future research.

\bibliographystyle{unsrt}
\bibliography{example_paper}

\clearpage
\appendix
\setcounter{equation}{0}

\section{Appendix}\label{apd}
\subsection{Detailed Derivation for Elimination Algorithm}\label{apd1}
Eliminating redundancy in GNN propagation is a complex challenge. In this section, we provide a detailed explanation of the thought process and derivation behind the elimination algorithm. During GNN propagation, node representations often redundantly aggregate features from previously visited nodes. This occurs because graphs inherently contain paths that allow nodes to be revisited, including undirected edges, backtracking edges, and self-loops. To address this issue, we introduce an elimination algorithm that filters out the redundant information introduced by these edges during the propagation process.

In contrast to the use of GAT in the main manuscript, we employ GCN here as an illustrative example, noting that the two are conceptually equivalent. The complete formula of GCN at the \(l\)-th layer is given by:
\begin{equation}\label{apd_eq1}
h_i^{\left(l\right)}=\sigma\Big(W^{(l)}\Big({C_{ii}h}_i^{\left(l-1\right)}+\sum_{j\in\mu\left(i\right)-i}{C_{ij}h}_j^{\left(l-1\right)}\Big)\Big),
\end{equation}
where $\sigma(\cdot)$ is the nonlinear function between layers, such as the composite function $\mathrm{Relu}(\cdot)$ that consists of the activation function Relu. $W^{(l)}$ is the learnable parameter matrix for the $l$-th layer, $C_{ij}$ is used to normalize the node features, such that $C_{ij}={d_i}^{-{1}/{2}}\ast{d_j}^{-{1}/{2}}$, and $d_i$ denotes the degree of node $i$, and $h_i^{\left(0\right)}=x_i^{\left(0\right)}$, and $\mu\left(i\right)$ is the set of neighborhood nodes of node  $i$.

Since $\sigma$ and $W$ almost always occur together, we simplify the notation by omitting the parentheses between $\sigma$ and $W$ in the following equations. The goal of the elimination algorithm is to eliminate the feature redundancy in the propagation process. There is no redundancy in the first layer, but since this formula will be used later, we still give it here:
\begin{equation}\label{apd_eq2}
h_i^{\left(1\right)}=\sigma W^{(1)}\Big({C_{ii}h}_i^{\left(0\right)}+\sum_{j\in\mu\left(i\right)-i}{C_{ij}h}_j^{\left(0\right)}\Big).
\end{equation}
The second layer is calculated as follows:
\begin{equation}\label{apd_eq3}
h_i^{\left(2\right)}=\sigma W^{(2)}\Big({C_{ii}h}_i^{\left(1\right)}+\sum_{j\in\mu\left(i\right)-i}{C_{ij}h}_j^{\left(1\right)}\Big).
\end{equation}
We try to rewrite the second term on the right side of the formula, with the goal of splitting the $h_i^{\left(0\right)}$ contained in it. At this time, there is: 
\begin{align}\label{apd_eq4}
\sum_{j\in\mu\left(i\right)-i}{C_{ij}h}_j^{\left(1\right)}=&\sum_{j\in\mu\left(i\right)-i}{C_{ij}\sigma W^{(1)}}\Big(C_{jj}h_j^{\left(0\right)}  + {C_{ji}h}_i^{\left(0\right)}+\sum_{k\in\mu\left(j\right)-j-i}{C_{jk}h}_k^{\left(0\right)}\Big),
\end{align}

Given that $\sigma(\cdot)$ is a nonlinear function, we usually cannot simplify this equation further. However, for the sake of discussion, we assume that $\sigma(\cdot)$ is $\mathrm{Relu}=\max\left(0,x\right)$, and that all the initial features $h^{\left(0\right)}$ have the same sign and all the parameters in the same column of $W$ have the same sign as well. Under these assumptions, we can split the terms inside the nonlinear function. Note that $C_{ij}$ is always non-negative, so we have:
\begin{footnotesize}
\begin{align}\label{apd_eq5}
\sum_{j\in\mu\left(i\right)-i}{C_{ij}h}_j^{\left(1\right)}&=\sum_{j\in\mu\left(i\right)-i}{C_{ij}W^{(1)}\sum_{k\in\mu\left(j\right)-j-i}{C_{jk}h_k^{\left(0\right)}}} +\sum_{j\in\mu\left(i\right)-i}{C_{ij}\Big|W^{(1)}\sum_{k\in\mu\left(j\right)-j-i}{C_{jk}h_k^{\left(0\right)}}\Big|}\nonumber\\
&+\sum_{j\in\mu\left(i\right)-i}{C_{ij}W^{(1)}\left(C_{jj}h_j^{\left(0\right)}
+{C_{ji}h}_i^{\left(0\right)}\right)}+\sum_{j\in\mu\left(i\right)-i}{C_{ij}\left(\left|W^{(1)}\left(C_{jj}h_j^{\left(0\right)}+{C_{ji}h}_i^{\left(0\right)}\right)\right|\right)},
\end{align}
\end{footnotesize}there $h_i^{\left(1\right)}$ in Eq.(\ref{apd_eq3}) already contains features around one hop, that the features about 0-hop and 1-hop in the second term are redundant, that is, we need to delete the third and fourth terms from the right side of the equation. For convenience, we denote the redundancy terms as:
\begin{equation}\label{apd_eq6}
{r}_i^{(1)}=\sum_{j\in\mu\left(i\right)-i}{C_{ij}\sigma W^{(1)}\left(C_{jj}h_j^{\left(0\right)}+C_{ji}h_i^{\left(0\right)}\right)},
\end{equation}
so that $h_i^{\left(2\right)}$ after elimination can be written as:
\begin{equation}\label{apd_eq7}
h_i^{\left(2\right)}=\sigma W^{(2)}\Big({C_{ii}h}_i^{\left(1\right)}-{r}_i^{(1)}+\sum_{j\in\mu\left(i\right)-i}{C_{ij}h}_j^{\left(1\right)}\Big).
\end{equation}
In addition, according to Eq.(\ref{apd_eq5}), we can also get another expression of this formula:
\begin{footnotesize}
\begin{equation}\label{apd_eq8}
h_i^{\left(2\right)}=\sigma{W^{(2)}}\Big({C_{ii}h}_i^{\left(1\right)}+\sum_{j\in\mu\left(i\right)-i}{C_{ij}\sigma W^{(1)}\sum_{k\in\mu\left(j\right)-j-i}{C_{jk}h}_k^{\left(0\right)}}\Big).
\end{equation}
\end{footnotesize}

Continue to consider the third layer. For clarity, we denote by \(\mathcal{T}^{(l)}(x) := \sigma\!\big(W^{(l)} x\big)\) the complete per-layer transformation, where \(x\) is the aggregated input from the previous layer. With the same premise, we first write out the formula of the third layer:
\begin{equation}\label{apd_eq9}
h_i^{\left(3\right)}=\mathcal{T}^{(3)}\Big({C_{ii}h}_i^{\left(2\right)}+\sum_{j\in\mu\left(i\right)-i}{C_{ij}h}_j^{\left(2\right)}\Big).
\end{equation}
According to Eq.(\ref{apd_eq8}), rewrite the second term in the above formula again:
\begin{footnotesize}
\begin{align}\label{apd_eq10}
\sum_{j\in\mu\left(i\right)-i}{C_{ij}h}_j^{\left(2\right)}&=\sum_{j\in\mu\left(i\right)-i}{C_{ij}\mathcal{T}^{(2)}\Big(C_{jj}h_j^{\left(1\right)}} +\sum_{k\in\mu\left(j\right)-j}{C_{jk}\mathcal{T}^{(1)}\sum_{l\in\mu\left(k\right)-k-j}{C_{kl}h_l^{\left(0\right)}}}\Big)\nonumber\\
&=\sum_{j\in\mu\left(i\right)-i}{C_{ij}\mathcal{T}^{(2)}\Big(C_{jj}h_j^{\left(1\right)}} +C_{ji}\mathcal{T}^{(1)}\sum_{l\in\mu\left(i\right)-i-j}{C_{il}h}_l^{\left(0\right)}\nonumber\\
&\qquad\qquad+\sum_{k\in\mu\left(j\right)-j-i}{C_{jk}\mathcal{T}^{(1)}\sum_{l\in\mu\left(k\right)-k-j}{C_{kl}h_l^{\left(0\right)}}}\Big),
\end{align}
\end{footnotesize}since the activation function Relu guarantees that each term has the same sign, there is:
\begin{scriptsize}
\begin{align}\label{apd_eq11}
\sum_{j\in\mu\left(i\right)-i}{C_{ij}h}_j^{\left(2\right)}&=\sum_{j\in\mu\left(i\right)-i}\bigg({C_{ij}W^{(2)}\bigg(C_{jj}h_j^{\left(1\right)}}+C_{ji}\mathcal{T}^{(1)}\sum_{l\in\mu\left(i\right)-i-j}{C_{il}h}_l^{\left(0\right)}\nonumber\\
&\qquad\qquad+\sum_{k\in\mu\left(j\right)-j-i}{C_{jk}\mathcal{T}^{(1)}\sum_{l\in\mu\left(k\right)-k-j}{C_{kl}h}_l^{\left(0\right)}}\bigg)\nonumber\\
&\qquad\qquad+C_{ij}\bigg(\bigg|W^{(2)}\Big(C_{jj}h_j^{\left(1\right)}+C_{ji}\mathcal{T}^{(1)}\sum_{l\in\mu\left(i\right)-i-j}{C_{il}h}_l^{\left(0\right)}\Big)\bigg|\nonumber\\
&\qquad\qquad+\bigg|W^{(2)}\Big(\sum_{k\in\mu\left(j\right)-j-i}{C_{jk}\mathcal{T}^{(1)}\sum_{l\in\mu\left(k\right)-k-j}{C_{kl}h}_l^{\left(0\right)}}\Big)\bigg|\bigg)\bigg).
\end{align}
\end{scriptsize}The information about two hops around the node in the above formula is unnecessary. It is important to note that the neighborhood of $i$ necessarily contains $i$ itself, whereas the neighborhood of $j$ contains both $i$ and $j$. However, the neighborhood of $k$ does not have this property, and $i$ may or may not be included in it. Therefore, we only need to delete $j$ and $k$ from the neighborhood of $k$. When $k$ and $i$ are connected by an edge, there is a loop in the graph, but we cannot know whether this loop exists or not. We can only assume that there is no loop here, and it is similar later. Therefore, \textbf{elimination only holds for acyclic graphs}. The redundancy terms at this time are: 
\begin{align}\label{apd_eq12}
{r}_i^{(2)}&=\sum_{j\in\mu\left(i\right)-i}{C_{ij}\bigg(W^{(2)}\Big(C_{jj}h_j^{\left(1\right)}+C_{ji}\mathcal{T}^{(1)}\sum_{l\in\mu\left(i\right)-i-j}{C_{il}h}_l^{\left(0\right)}\Big)} \nonumber \\
&\qquad\qquad+\bigg|W^{(2)}\Big(C_{jj}h_j^{\left(1\right)}
+C_{ji}\mathcal{T}^{(1)}\sum_{l\in\mu\left(i\right)-i-j}{C_{il}h}_l^{\left(0\right)}\Big)\bigg|\bigg)\nonumber \\
&=\sum_{j\in\mu\left(i\right)-i}{C_{ij}\mathcal{T}^{(2)}\bigg(C_{jj}h_j^{\left(1\right)}}+C_{ji}\mathcal{T}^{(1)}\sum_{l\in\mu\left(i\right)-i-j}{C_{il}h}_l^{\left(0\right)}\bigg),
\end{align}
where $h_l^{(0)}$ is essentially a one-hop neighborhood node of $i$. We transform Eq.(\ref{apd_eq2}) as follows:
\begin{equation}\label{apd_eq13}
h_i^{\left(1\right)}=\mathcal{T}^{(1)}\Big({C_{ii}h}_i^{\left(0\right)}+{C_{ij}h}_j^{\left(0\right)}+\sum_{l\in\mu\left(i\right)-i-j}{C_{il}h}_l^{\left(0\right)}\Big),
\end{equation}
so, there is:

\begin{align}\label{apd_eq14}
C_{ji}\mathcal{T}^{(1)}\sum_{l\in\mu\left(i\right)-i-j}{C_{il}h_l^{\left(0\right)}}=C_{ji}h_i^{\left(1\right)}-C_{ji}\mathcal{T}^{(1)}\left({C_{ii}h}_i^{\left(0\right)}+{C_{ij}h}_j^{\left(0\right)}\right).
\end{align}
Substituting Eq.(\ref{apd_eq14}) into Eq.(\ref{apd_eq12}), we have:
\begin{align}\label{apd_eq15}
{r}_i^{(2)}=\sum_{j\in\mu\left(i\right)-i}{C_{ij}\mathcal{T}^{(2)}\Big(C_{jj}h_j^{\left(1\right)}+C_{ji}h_i^{\left(1\right)}} -C_{ji}\mathcal{T}^{(1)}\big({C_{ii}h}_i^{\left(0\right)}+{C_{ij}h}_j^{\left(0\right)}\big)\Big),
\end{align}
so that $h_i^{\left(3\right)}$ after elimination can be written as:
\begin{align}\label{apd_eq16}
h_i^{\left(3\right)}=\mathcal{T}^{(3)}\Big({C_{ii}h}_i^{\left(2\right)}-{r}_i^{(2)}+\sum_{j\in\mu\left(i\right)-i}{C_{ij}h}_j^{\left(2\right)}\Big).
\end{align}
According to Eq.(\ref{apd_eq11}), $h_i^{(3)}$ can also be written as:
\begin{footnotesize}
\begin{align}\label{apd_eq17}
h_i^{\left(3\right)}=\mathcal{T}^{(3)}\Big({C_{ii}h}_i^{\left(2\right)}+\sum_{j\in\mu\left(i\right)-i}{C_{ij}\mathcal{T}^{(2)}\sum_{k\in\mu\left(j\right)-j-i}{C_{jk}\mathcal{T}^{(1)}\sum_{l\in\mu\left(k\right)-k-j}{C_{kl}h}_l^{\left(0\right)}}}\Big).
\end{align}
\end{footnotesize}

Continue to consider the fourth layer, and the corresponding equation is:
\begin{align}\label{apd_eq18}
h_i^{\left(4\right)}=\mathcal{T}^{(4)}\Big({C_{ii}h}_i^{\left(3\right)}+\sum_{j\in\mu\left(i\right)-i}{C_{ij}h}_j^{\left(3\right)}\Big).
\end{align}
Rewrite the above formula using Eq.(\ref{apd_eq17}). According to previous discussions, the items of f can be split directly at this point. For simplicity, we directly give the result after splitting:
\begin{scriptsize}
\begin{align}\label{apd_eq19}
\sum_{j\in\mu\left(i\right)-i}{C_{ij}h}_j^{\left(3\right)}&=\sum_{j\in\mu\left(i\right)-i}{C_{ij}\mathcal{T}^{(3)}\bigg({C_{jj}h}_j^{\left(2\right)}} +\sum_{k\in\mu\left(j\right)-j}{C_{jk}\mathcal{T}^{(2)}\sum_{l\in\mu\left(k\right)-k-j}{C_{kl}\mathcal{T}^{(1)}\sum_{m\in\ \ \mu\left(l\right)-l-k}{C_{lm}h}_{m}^{\left(0\right)}}}\bigg)\nonumber\\
&=\sum_{j\in\mu\left(i\right)-i}{C_{ij}\mathcal{T}^{(3)}\bigg({C_{jj}h}_j^{\left(2\right)}}+C_{ji}\mathcal{T}^{(2)}\sum_{l\in\mu\left(i\right)-i-j}{C_{il}\mathcal{T}^{(1)}\sum_{m\in\ \ \mu\left(l\right)-l-i}{C_{lm}h}_m^{\left(0\right)}}\bigg)\nonumber\\
&\quad\quad+\sum_{j\in\mu\left(i\right)-i}{C_{ij}\mathcal{T}^{(3)}\sum_{k\in\mu\left(j\right)-j-i} {C_{jk}\mathcal{T}^{(2)}}}\sum_{l\in\mu\left(k\right)-k-j}{C_{kl} \mathcal{T}^{(1)}\sum_{m\in\ \ \mu\left(l\right)-l-k}{C_{lm}h}_{m}^{\left(0\right)}}.
\end{align}
\end{scriptsize}
Transform Eq.(\ref{apd_eq8}) as follows:
\begin{footnotesize}
\begin{align}\label{apd_eq20}
h_i^{\left(2\right)}&=\mathcal{T}^{(2)}\Big({C_{ii}h}_i^{\left(1\right)}+C_{ij}\mathcal{T}^{(1)}\sum_{k\in\mu\left(j\right)-j-i}{C_{jk}h}_k^{\left(0\right)}+\sum_{l\in\mu\left(i\right)-i-j}{C_{il}\mathcal{T}^{(1)}\sum_{m\in\mu\left(l\right)-l-i}{C_{lm}h}_m^{\left(0\right)}}\Big),
\end{align}
\end{footnotesize}
so, there is:
\begin{scriptsize}
\begin{align}\label{apd_eq21}
C_{ji}\mathcal{T}^{(2)}&\sum_{l\in\mu\left(i\right)-i-j}{C_{il}\mathcal{T}^{(1)}\sum_{m\in\mu\left(l\right)-l-i}{C_{lm}h}_m^{\left(0\right)}}=C_{ji}h_i^{\left(2\right)}-\mathcal{T}^{(2)}\Big({C_{ii}h}_i^{\left(1\right)}+C_{ij}\mathcal{T}^{(1)}\sum_{k\in\mu\left(j\right)-j-i}{C_{jk}h}_k^{\left(0\right)}\Big).
\end{align}
\end{scriptsize}Substituting Eq.(\ref{apd_eq21}) into Eq.(\ref{apd_eq19}), we get that this layer’s redundancy term is:
\begin{footnotesize}
\begin{align}\label{apd_eq22}
{r}_i^{(3)}&=\sum_{j\in\mu\left(i\right)-i}{C_{ij}\mathcal{T}^{(3)}\Big({C_{jj}h}_j^{\left(2\right)}+C_{ji}h_i^{\left(2\right)}}  -\mathcal{T}^{(2)}\big({C_{ii}h}_i^{\left(1\right)}+C_{ij}\mathcal{T}^{(1)}\sum_{k\in\ \ \mu\left(j\right)-j-i}{C_{jk}h}_k^{\left(0\right)}\big)\Big).
\end{align}
\end{footnotesize}According to Eq.(\ref{apd_eq2}), there is such an equation:
\begin{align}\label{apd_eq23}
h_j^{\left(1\right)}=\mathcal{T}^{(1)}\Big({C_{jj}h}_j^{\left(0\right)}+{C_{ji}h}_i^{\left(0\right)}+\sum_{k\in\mu\left(j\right)-j-i}{C_{jk}h}_k^{\left(0\right)}\Big).
\end{align}
The acquisition method of Eq.(\ref{apd_eq23}) is consistent with that of Eq.(\ref{apd_eq13}). The difference is only to replace node $i$ with node $j$. The subscript of the final term uses $j$ or $l$ to indicate that there is no difference at this point. According to Eq.(\ref{apd_eq23}), the redundancy term becomes:
\begin{scriptsize}
\begin{align}\label{apd_eq24}
{r}_i^{(3)}&=\sum_{j\in\mu\left(i\right)-i}{C_{ij}\mathcal{T}^{(3)}\bigg({C_{jj}h}_j^{\left(2\right)}+C_{ji}h_i^{\left(2\right)}}-\mathcal{T}^{(2)}\Big({C_{ii}h}_i^{\left(1\right)}+C_{ij}h_j^{\left(1\right)}-C_{ij}\mathcal{T}^{(1)}\big({C_{jj}h}_j^{\left(0\right)}+{C_{ji}h}_i^{\left(0\right)}\big)\Big)\bigg).
\end{align}
\end{scriptsize}so, there is:
\begin{align}\label{apd_eq25}
h_i^{\left(4\right)}=\mathcal{T}^{(4)}\Big({C_{ii}h}_i^{\left(3\right)}-{r}_i^{(3)}+\sum_{j\in\mu\left(i\right)-i}{C_{ij}h}_j^{\left(3\right)}\Big),
\end{align}
and there is:
\begin{footnotesize}
\begin{align}
\label{apd_eq26}
h_i^{(4)}
&= \mathcal{T}^{(4)}\Big(
    C_{ii} h_i^{(3)}
    + \sum_{j\in \mu(i)-i}
      C_{ij}\,\mathcal{T}^{(3)}
      \sum_{k\in \mu(j)-j-i}
      C_{jk}\,\mathcal{T}^{(2)} 
      \sum_{l\in \mu(k)-k-j}
      C_{kl}\,\mathcal{T}^{(1)}
      \sum_{m\in \mu(l)-l-k}
      C_{lm}\, h_m^{(0)}
\Big).
\end{align}
\end{footnotesize}

By analogy, comparing Eq.(\ref{apd_eq24}), Eq.(\ref{apd_eq15}) and Eq.(\ref{apd_eq6}), it can be found that any layer’s redundancy term $r$ can always be simplified to a form expressed by $h_i^{\left(0\right)}$ and $h_j^{\left(0\right)}$. In order to accurately describe this redundancy term, we first define a nested function as follows:
\begin{footnotesize}
\begin{align}\label{apd_eq27}
f_{ij}^{(1)}&=C_{ij}\mathcal{T}^{(1)}\left(C_{jj}h_j^{\left(0\right)}+C_{ji}h_i^{\left(0\right)}\right), \nonumber \\
f_{ij}^{(2)}&=C_{ij}\mathcal{T}^{(2)}\Big({C_{jj}h}_j^{\left(1\right)}+C_{ji}h_i^{\left(1\right)}-C_{ji}\mathcal{T}^{(1)}\left({C_{ii}h}_i^{\left(0\right)}+{C_{ij}h}_j^{\left(0\right)}\right)\Big),\nonumber \\
&\cdots\nonumber \\
f_{ij}^{(l)}&=C_{ij}\mathcal{T}^{(l)}\Big({C_{jj}h}_j^{\left(l-1\right)}+C_{ji}h_i^{\left(l-1\right)}-C_{ji}\mathcal{T}^{(l-1)}\Big({C_{ii}h}_i^{\left(l-2\right)}+{C_{ij}h}_j^{\left(l-2\right)}- f_{ji}^{(l-2)}\Big)\Big),
\end{align}
\end{footnotesize}
where $l>2$. At this time, the redundancy term of layer $l+1$ can be expressed as: 
\begin{align}\label{apd_eq28}
{r}_i^{(l)}=\sum_{j\in\mu\left(i\right)-i} f_{ij}^{(l)}.
\end{align}
So that GCN eliminates computing generalization at layer $l+1$ as follows:

\begin{equation}\label{apd_eq29}
h_i^{\left(l+1\right)}=\mathcal{T}^{(l+1)}\Big({C_{ii}h}_i^{\left(l\right)}-{r}_i^{(l)}+\sum_{j\in\mu\left(i\right)-i}{C_{ij}h}_j^{\left(l\right)}\Big).
\end{equation}
This result can be verified by experiment, and the result after eliminating redundancy is shown in Figure \ref{apd:fig1}.


While the graph elimination algorithm effectively removes feature redundancy in neighborhood aggregation and disentangles representations at each hop, its computational cost can become prohibitive on large-scale graphs and presents challenges when dealing with sparse tensors. To address this issue, we propose a simpler and more efficient alternative: self-loop elimination. By removing self-loops from the GNN layers, the receptive field ensures that only nodes at odd or even hop distances are included in the outcomes of odd or even propagation steps, respectively. This strategy splits odd- and even-hop features without introducing additional costs, thereby enhancing hop-level attention. In many cases, self-loop elimination offers a more economical choice for balancing efficiency and performance. Nevertheless, it is important to note that, similar to the graph elimination algorithm, self-loop elimination is applicable only to acyclic graphs.

\subsection{Implementation Details}
\label{apd:expde}
\textbf{Datasets and Tasks}\quad
We evaluate our model on five public benchmarks:  
(1) \textbf{LRGB} \cite{ref72};  
(2) \textbf{OGB} \cite{ref34};  
(3) classical node‑classification datasets—Cora, CiteSeer, PubMed \cite{ref98}, Photo \cite{ref99}, and WikiCS \cite{ref100};  
(4) ZINC \cite{zinc} (added in this appendix);  
(5) five heterophilous‑graph tasks \cite{ref92}: Squirrel, Chameleon, Amazon\_Ratings, Roman\_Empire, and Questions (also added here).

\textbf{Data Splits and Evaluation Protocol}\quad
For benchmarks with official splits—OGB, and LRGB—we strictly follow the provided train/validation/test partitions.  
For the node‑classification and heterophilous‑graph tasks we reproduce the splits released by \cite{ref111} in their \href{https://github.com/LUOyk1999/tunedGNN}{public repository}.  
All results are reported on the test set at the epoch where the validation performance peaks.  
Repetition counts are: five runs on LRGB, ten runs on OGB, and 3–5 runs on the node‑classification and heterophilous‑graph tasks, matching Luo \textit{et al.}

\textbf{Hyper‑parameter Search}\quad
For node-classification and heterophilous-graph tasks, we perform Bayesian optimisation using \textsc{Optuna} \cite{ref110}. The search space is defined as follows:  
$L \in \{2, \dots, 12\}$; learning rate $\eta \in [10^{-4}, 10^{-2}]$ (log-uniform);  
weight decay $w_d \in \{10^{-1}, 10^{-2}, 10^{-3}, 5 \times 10^{-4}, 10^{-4}, 5 \times 10^{-5}, 10^{-5}, 10^{-6}\}$;  
dropout $p \in [0, 0.9]$ with an interval of 0.1;  
batch size and hidden dimension are selected from $\{2^4, 2^5, \dots, 2^{11}\}$;  
number of attention heads $h \in \{1, \dots, 10\}$;  
training epochs $\in \{200, 300, 500, 1000, 1500, 2000, 2500\}$;  
propagation steps $K \in \{1, \dots, 8\}$;  
compression coefficient $\gamma \in [0, 1]$.  
For the remaining tasks, we start from the default GCN configuration and manually fine-tune within the same ranges.  
All optimal configurations are available in our \href{https://github.com/tmp488598/Graph-Elimination-Networks}{repository}.

\begin{table}[t]
\centering
\caption{Node classification results over homophilous graphs in small-scale real-world scenarios (\%). Comparison of GENs and baseline methods, with the best result highlighted in bold, second and third best results underlined.}
\label{tab2}
\resizebox{0.95\textwidth}{!}{%
\begin{tabular}{lcccccc}
\toprule
 \textbf{Dataset} & \textbf{Cora} & \textbf{CiteSeer} & \textbf{PubMed} & \textbf{Photo} & \textbf{WikiCS} \\
\midrule
\# nodes       & 2,708 & 3,327 & 19,717 & 7,650 & 11,701 \\
\# edges       & 5,278 & 4,732 & 44,324 & 119,081 & 216,123 \\
Metric         & Accuracy ↑ & Accuracy ↑ & Accuracy ↑ & Accuracy ↑ & Accuracy ↑  \\
\midrule
\textbf{GCN}         & \underline{85.10 $\pm$ 0.67} & \underline{73.14 $\pm$ 0.50} &  \underline{81.12 $\pm$ 0.22} & 96.10 $\pm$ 0.46 & 80.30 $\pm$ 0.62 \\
\textbf{GraphSAGE}   & 83.88 $\pm$ 0.65 & 72.26 $\pm$ 0.55 & 79.72 $\pm$ 0.50 & \underline{96.78 $\pm$ 0.23} &  \underline{80.69 $\pm$ 0.31} \\
\textbf{GAT}         & 84.46 $\pm$ 0.55 & 72.22 $\pm$ 0.84 & 80.28 $\pm$ 0.12 &  \underline{96.60 $\pm$ 0.33} & \underline{81.07 $\pm$ 0.54} \\
\midrule
\textbf{GraphGPS}    & 83.87 $\pm$ 0.96 & 72.73 $\pm$ 1.23 & 79.94 $\pm$ 0.26 & 94.89 $\pm$ 0.14 & 78.66 $\pm$ 0.49 \\
\textbf{NAGphormer}  & 80.92 $\pm$ 1.17 & 70.59 $\pm$ 0.89 & 80.14 $\pm$ 1.06 & 96.14 $\pm$ 0.16 & 77.92 $\pm$ 0.93 \\
\textbf{Exphormer}   & 83.29 $\pm$ 1.36 & 71.85 $\pm$ 1.11 & 79.67 $\pm$ 0.73 & 95.69 $\pm$ 0.39 & 79.38 $\pm$ 0.82 \\
\textbf{GOAT}        & 83.26 $\pm$ 1.24 & 72.21 $\pm$ 1.29 & 80.06 $\pm$ 0.67 & 94.33 $\pm$ 0.21 & 77.96 $\pm$ 0.03 \\
\textbf{NodeFormer}  & 82.73 $\pm$ 0.75 & 72.37 $\pm$ 1.20 & 79.59 $\pm$ 0.92 & 93.43 $\pm$ 0.56 & 75.13 $\pm$ 0.93 \\
\textbf{SGFormer}    & \underline{84.82 $\pm$ 0.85} & \underline{72.72 $\pm$ 1.15} &  \underline{80.60 $\pm$ 0.49} & 95.58 $\pm$ 0.36 & 80.05 $\pm$ 0.46 \\
\textbf{Polynormer}  & 83.43 $\pm$ 0.90 & 72.19 $\pm$ 0.83 & 79.35 $\pm$ 0.73 & 96.57 $\pm$ 0.23 & 80.26 $\pm$ 0.92 \\
\midrule
\textbf{GENs}& \textbf{85.43 $\pm$ 0.42} &\textbf{73.56 $\pm$ 0.46}&\textbf{81.94 $\pm$ 0.34} &\textbf{97.12 $\pm$ 0.24}&\textbf{81.48 $\pm$ 0.46}\\
\bottomrule
\end{tabular}
}
\end{table}

\textbf{Experimental Environment}\quad
All experiments are conducted on a single NVIDIA A100‑80GB GPU, and no early‑stopping strategy is employed. Evaluation metrics and dataset statistics are included in their respective tables.

\begin{table}[b]
\centering
\caption{Node classification results on heterophilous graphs (\%). The best result is highlighted in bold, and the second and third best results are underlined.}
\label{apd_tab1}
\resizebox{\textwidth}{!}{%
\begin{tabular}{lcccccc}
\toprule
 \textbf{Dataset} & \textbf{Squirrel} & \textbf{Chameleon} & \textbf{Amazon-Ratings} & \textbf{Roman-Empire} & \textbf{Questions} \\
\midrule
\# nodes       & 2,223 & 890 & 24,492 & 22,662 & 48,921 \\
\# edges       & 46,998 & 8,854 & 93,050 & 32,927 & 153,540 \\
Metric         & Accuracy ↑ & Accuracy ↑ & Accuracy ↑ & Accuracy ↑ & ROC-AUC ↑ \\
\midrule
\textbf{GCN}        & \underline{45.01 $\pm$ 1.63} & \underline{46.29 $\pm$ 3.40} & 53.80 $\pm$ 0.00 &\underline{91.27 $\pm$ 0.20} & \underline{79.02 $\pm$ 0.60} \\
\textbf{GraphSAGE}  & 40.78 $\pm$ 1.47 & 44.81 $\pm$ 4.74 & \underline{55.40 $\pm$ 0.21} & 91.06 $\pm$ 0.27 & 77.21 $\pm$ 1.28 \\
\textbf{GAT}        & 41.73 $\pm$ 2.07 & 44.13 $\pm$ 2.41 & 54.92 $\pm$ 0.61 & 90.63 $\pm$ 0.14 & 77.95 $\pm$ 0.51 \\
\midrule
\textbf{H2GCN}  & 35.10 $\pm$ 1.15 & 26.75 $\pm$ 3.64 & 36.47 $\pm$ 0.23 & 60.11 $\pm$ 0.52 & 63.59 $\pm$ 1.46 \\
\textbf{CPGNN}  & 30.04 $\pm$ 2.03 & 33.00 $\pm$ 3.15 & 39.79 $\pm$ 0.77 & 63.96 $\pm$ 0.62 & 65.96 $\pm$ 1.95 \\
\textbf{GPRGNN} & 38.95 $\pm$ 1.99 & 39.93 $\pm$ 3.30 & 44.88 $\pm$ 0.34 & 64.85 $\pm$ 0.27 & 55.48 $\pm$ 0.91 \\
\textbf{FSGNN}  & 35.92 $\pm$ 1.32 & 40.61 $\pm$ 2.97 & 52.74 $\pm$ 0.83 & 79.92 $\pm$ 0.56 & 78.86 $\pm$ 2.92 \\
\textbf{GloGNN} & 35.11 $\pm$ 1.24 & 25.90 $\pm$ 3.58 & 36.89 $\pm$ 0.14 & 59.63 $\pm$ 0.69 & 65.74 $\pm$ 1.19 \\
\midrule
\textbf{GraphGPS}   & 39.81 $\pm$ 2.28 & 41.55 $\pm$ 3.91 &53.27 $\pm$ 0.06 & 82.72 $\pm$ 0.08 & 72.56 $\pm$ 1.33 \\
\textbf{NodeFormer} & 38.89 $\pm$ 2.67 & 36.38 $\pm$ 3.85 & 43.79 $\pm$ 2.57 & 74.83 $\pm$ 0.81 & 75.02 $\pm$ 1.61 \\
\textbf{SGFormer}   & \underline{42.65 $\pm$ 2.41} & \underline{45.21 $\pm$ 3.72} & 54.14 $\pm$ 2.00 & 80.01 $\pm$ 0.44 & 73.81 $\pm$ 0.59 \\
\textbf{Polynormer} & 41.97 $\pm$ 2.14 & 41.97 $\pm$ 3.18 & \underline{54.96 $\pm$ 2.22} & \textbf{92.66 $\pm$ 0.60} & \underline{78.94 $\pm$ 0.78} \\
\midrule
\textbf{GENs}& \textbf{46.25 $\pm$ 0.61} & \textbf{47.10 $\pm$ 4.40} & \textbf{55.92 $\pm$ 0.43} & \underline{92.52 $\pm$ 0.50} & \textbf{79.22$\pm$ 0.74}\\
\bottomrule
\end{tabular}
}
\end{table}

\begin{table}[t]
\centering
\caption{Comparison of Mean Absolute Error (MAE) for GNNs on Acyclic and Cyclic Subsets of the ZINC Dataset.}
\label{apd_tab3}
\resizebox{\textwidth}{!}{%
\begin{tabular}{l c ccccc}
\toprule
\textbf{Dataset} & \textbf{\#Samples} & \textbf{GENs} & \textbf{GAT} & \textbf{GCN}  & \textbf{GraphGPS} & \textbf{Exphormer} \\
\midrule
ZINC (Acyclic) & 1,109 & \textbf{0.095} $\pm$ 0.011 & 0.151 $\pm$ 0.016 & 0.141 $\pm$ 0.012 & 0.142 $\pm$ 0.018 & 0.174 $\pm$ 0.015 \\
ZINC (Cyclic) & 12K & 0.074 $\pm$ 0.012 & 0.384 $\pm$ 0.007 & 0.367 $\pm$ 0.011  & \textbf{0.070} $\pm$ 0.004 & 0.132 $\pm$ 0.005 \\
\midrule
\textbf{Abs. Diff.} & - & -0.021 & +0.233 & +0.226 & -0.072 & -0.042 \\
\bottomrule
\end{tabular}
}
\end{table}

\subsection{Additional Results}\label{app:exp_add}

In this section, we report results for two additional tasks excluded from the main text due to space limitations: heterophilous-graph tasks and performance comparisons on acyclic graphs.

We further evaluated GENs in small- to medium-scale real-world scenarios, with results shown in Table \ref{tab2}. Contrary to expectations from prior reports, advanced GT methods like SGFormer did not outperform well-tuned traditional GNNs \cite{ref92}. GENs achieved a modest improvement of around 0.5–1\% over GCN, yet consistently delivered the best performance across all tasks. This consistent advantage underscores the practical effectiveness of GENs and suggests that, despite approximating GTs, they are highly competitive with traditional GNNs in small- to medium-scale settings, demonstrating both robustness and versatility in real-world graph applications.

Full results for the heterophilous graph tasks are detailed in Table \ref{apd_tab1}. On small-scale heterophilous graphs, GENs surpass state-of-the-art Graph Transformers and consistently outperform traditional methods across all tasks. These results align with the findings from the small- to medium-scale experiments, further highlighting the versatility and robustness of GENs across diverse graph-based tasks.

To examine the influence of cycles on GENs, we constructed an acyclic subset by extracting all cycle-free samples from the ZINC 250K dataset, resulting in 1,109 samples. This subset was partitioned following the original split: 950 for training, 32 for testing, and 127 for validation. We compared these results with a subset of the original ZINC dataset containing 12,000 samples, of which only 66 (55 training, 10 testing, and 1 validation) are acyclic, making it a predominantly cyclic collection. The outcomes are presented in Table \ref{apd_tab3}.
Results show that both GENs and GPS experienced performance degradation on the acyclic subset relative to the original dataset, whereas traditional GNNs (GCN, GAT) generally exhibited improved performance. This observation suggests that GENs are potentially more robust to the presence of cycles than traditional GNNs, and less dependent on dataset scale compared to GTs, further underscoring GENs’ performance advantages.

\end{document}